\documentclass[letterpaper, 10pt, conference]{ieeeconf}
\usepackage{array}
\newcolumntype{M}{>{$\vcenter\bgroup\hbox\bgroup}c<{\egroup\egroup$}}

\IEEEoverridecommandlockouts
\usepackage{amsmath}
\usepackage{amssymb}
\usepackage{booktabs}
\usepackage{siunitx}
\usepackage{graphicx}
\usepackage{algorithm}
\usepackage{color}
\usepackage{array}
\usepackage{url}
\usepackage{flushend}
\makeatletter
\let\NAT@parse\undefined
\makeatother
\usepackage[square, numbers]{natbib}

\usepackage{bm}
\usepackage{pseudocode}
\usepackage[noend]{algpseudocode}
\usepackage{graphicx}
\usepackage{hyperref}
\usepackage{tabularx}

\def\eqref#1{Eq.~\ref{#1}}

\newcommand{\be}{\mathbf{e}}
\newcommand{\bx}{\mathbf{x}}
\newcommand{\bz}{\mathbf{z}}
\newcommand{\bu}{\mathbf{u}}
\newcommand{\bq}{\mathbf{q}}

\newcommand{\SE}[1]{\mathit{SE}(#1)}

\title{\bf\LARGE Monte Carlo Localization in Hand-Drawn Maps}

\author{Bahram Behzadian\and Pratik Agarwal \and Wolfram Burgard \and Gian Diego Tipaldi %
  \thanks{\scriptsize All authors are with the University of Freiburg,
    Institute of Computer Science, 79110 Freiburg, Germany. This work has partly
    been supported by the European Commission under FP7-610532-SQUIRREL. }}

\begin{document}
\maketitle

\begin{abstract}
Robot localization is a one of the most important problems in robotics.
Most of the existing approaches assume that the map of the environment is available beforehand and focus on accurate metrical localization.
In this paper, we address the localization problem when the map of the environment is not present beforehand, and the robot relies on a hand-drawn map from a non-expert user.
We addressed this problem by expressing the robot pose in the pixel coordinate and simultaneously estimate a local deformation of the hand-drawn map.
Experiments show that we are able to localize the robot in the correct room with a robustness up to 80\%.

\end{abstract}

\section{Introduction}

Localization, the problem of estimating a robot pose in the environment, is probably one of the most studied and understood problems in mobile robotics.
Several solutions have been presented in the last decades, most of them based on probabilistic inference over the space of possible robot configurations~\cite{thrun2005probabilistic}.
Although the existing approaches have been demonstrated to be robust, efficient and very accurate~\cite{dellaert1999icra,fox03ijrr,sprunk2012iros}, they mostly rely on one major assumption: The existence of an already available map of the environment, built beforehand with the same sensing modality of the robot.

In some circumstances, however, building such a map could be a nuisance for the user.
This is the case, for instance, of vacuum cleaners, lawn mowers, pool cleaners and many other service robots.
Often, when the user buys such a robot, he or she wants to use it immediately without waiting for an expensive and time-consuming mapping routine to complete.
In some other cases, building an a-priori map is not even possible, for instance in environments that may harm humans, e.g., a mine field or a toxic factory.
Moreover, mapping algorithms may result in local minima and the resulting maps might be unusable for navigation.
Although automatic tools to detect such inconsistencies exist~\cite{mazuran2014icra}, they require an expert user to analyze the data to correct the map.

In this paper, we address the localization problem when no map of the environment is available beforehand.
We propose an algorithm that solely requires a hand-drawn map of the environment, sketched by the user.
We believe that drawing such a map puts no burden on the user and is an intuitive task.
Furthermore, we do not assume that the map is metrically accurate nor proportional up to a single scale.
Objects might be missing, and the deformation of the map might change at different locations.
This reduces the ability to accurately localize the robot in a metric sense, since the map is \emph{bended} in different ways and distances are not respected.
To address these problems, we extend the Monte Carlo localization algorithm in two ways.
First, we express the robot pose in the pixel coordinate of the drawing.
This resolves the metrical issues and provides a sort of normalization with respect to the deformation.
Second, we augment the state space of the robot with a deformation parameter and track the local deformation of the drawing over time.

\begin{figure}[t]
\begin{center}
\includegraphics[width=0.48\textwidth]{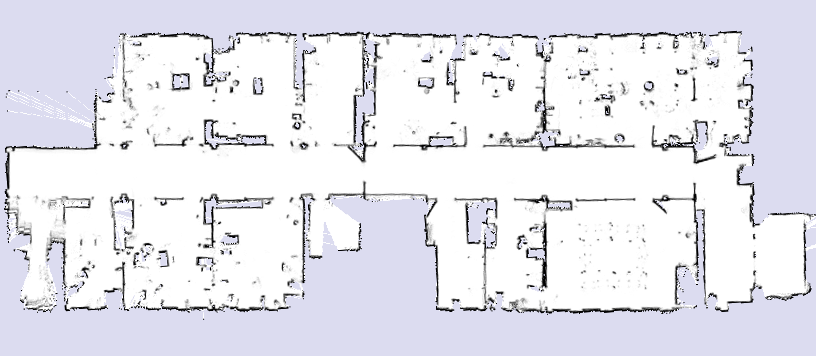}
\includegraphics[width=0.48\textwidth]{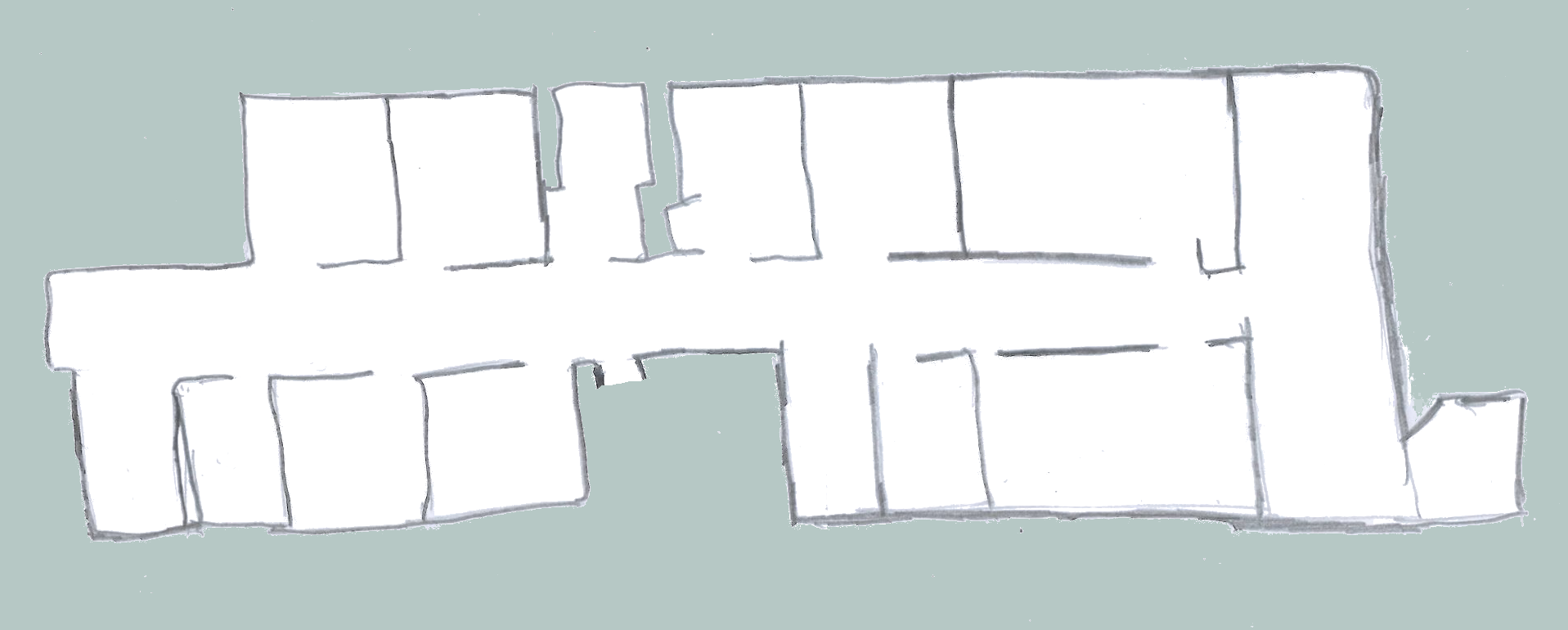}
\includegraphics[width=0.48\textwidth]{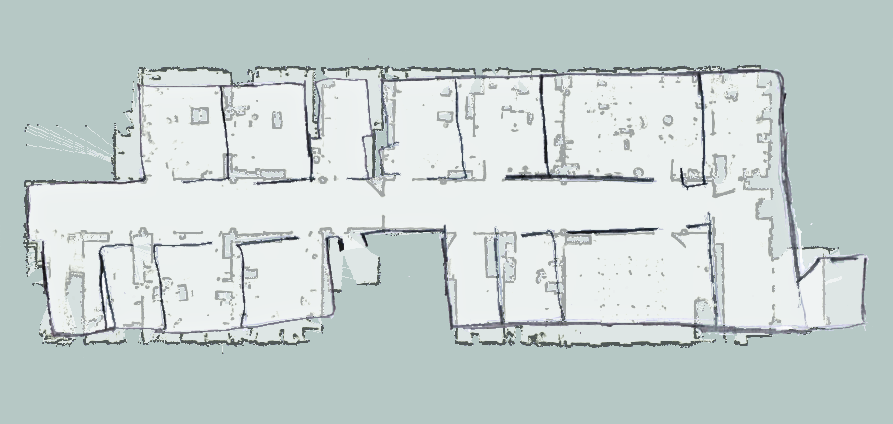}
\end{center}
  \caption{Occupancy grid of the dataset FR079 (top), built using a SLAM algorithm, an hand-drawn map used to localize the robot (middle) and their overlay (bottom).}
\label{fig:fr079}
\end{figure}

\section{Related work}

Robot localization is a widely studied problem~\cite{thrun2005probabilistic} and several approaches have been proposed in the past, such as Markov localization~\cite{fox99jair}, Monte Carlo localization (MCL)~\cite{dellaert1999icra}, and multiple hypothesis tracking (MHT)~\cite{jensfelt2001tra}.
Those approaches rely on the existence of a prior map and a static environment. Some researchers extended those approaches to be able to handle changes in the environment over time~\cite{tipaldi2013ijrr,krajnik2014ars,biswas2014icra}. However, they still rely on metrical maps and some prior information of the the environment to be built beforehand.

Few works have been done with respect to localization without prior maps.
\citet{koenig1996passive} propose a localization approach, where the user provides only a topological sketch of the environment.
Some authors used floor plan maps available from construction blueprints.
\citet{ito14icra} propose a localization algorithm that relies on blueprints of the environment.
They employ an RGB-D sensor to estimate the walls of the environment and match them to the floor plan.
To improve the initialization, they rely on a map of WiFi signal strengths.
\citet{setalaphruk2003cira} employ a multi-hypothesis filter on blueprints. They first compute a Voronoi diagram of the floor plan and store its intersections as landmarks. Second, they use the Voronoi intersection from the current readings and match them to the one computed from the map.
Contrary to our approach, they assume that the blueprint of the environment is metrically correct, and its scale is known.

Most of the works using hand-drawn maps exploit them as a mean of communication between the robot and a human operator.
\citet{kawamura2002ijra} present an approach where an operator first sketches a rough map of the environment with a set of way points to follow. They use a library of known objects to associate perceptions to the sketch map and triangulate the robot position using angle measurements.
\citet{skubic2007ar} Propose a sketch interface for guiding a robot. The user draws a sketch of both the environment and instructs the robot to go to certain locations. The system computes a path in the hand-drawn map, and the robot execute it in an open-loop controller, without localizing the robot.
\citet{shah2010iros} propose a similar approach. The focus of the work, however, is on how to extract qualitative instructions for the robot. They translate the instructions to robot commands and localize the robot using an overhead camera system.
\citet{yun2008iros} proposed and evaluated a quantitative measure of navigability on hand-drawn maps. The work only focuses on qualitative aspects of navigability but does not address the ability to localize in them. Moreover, the sketch maps that they considered are made of line segments and are automatically generated.
\citet{skubic2002aaais} propose an approach to qualitative reason about sketch maps. The authors are able to extract information such as an obstacle is on the right of the robot and give qualitative commands as turn left or go straight. The work has been extended by \citet{chronis2004icra} with the use of reactive controllers to guide the robot.
\citet{forbus2004aim} developed nuSketch, a qualitative reasoning framework exploiting topological properties and Voronoi diagrams. They provide answers to query like finding places with certain properties or properties of paths.
Our work is orthogonal to them, and the proposed localization algorithm can be integrated with any of those control interfaces.

Localization using hand-drawn maps has received very little attention from the robotics community.
\citet{parekh2007icra} presented a technique to perform scene matching between a map of the environment and a sketch using spatial relations.
They assume a set of objects is present in both the sketch and the map with known segmentation. They then use particle swarm optimization (PSO) techniques to compute the alignment and the sketch deformation.
\citet{matsuo2012iros} extended the previous work in a simultaneous localization and mapping (SLAM) framework. They assume, however that the sketch map is composed of rectangles corresponding to building in the scene.
\citet{shah2013ijrr} present an algorithm for controlling a mobile robot using a qualitative map consisting of landmarks and path waypoints. They assume that the both the sketch map and the real environment is made of point landmarks and also assume known data associations between them.
In contrast to them, our approach does not make any assumption on the format of the sketch map and treats it as a raster image.
Moreover, we do not attempt to \emph{transform} the hand-drawn map to reflect the real world but we directly localize the robot in the hand-drawn map.
To the best of our knowledge, the proposed approach is the first attempt to localize a robot from a generic hand-drawn map with no further assumptions.

\section{Localization in Hand-Drawn Maps}
In this section, we describe the extension we made in the original Monte Carlo localization algorithm~\cite{dellaert1999icra} for localizing in hand-drawn maps.
We propose two main extensions. First, we augment the state space of the robot with an additional variable that represents the local deformation of the map.
Second, we localize the robot in the pixel coordinate frame of the map, instead of the world coordinate frame. In order to do so, we extended both the motion and the observation model to be locally projected onto the hand-drawn map.

\subsection{Monte-Carlo Localization in Pixel Coordinates}

The goal of our work is to estimate the pose of the robot $\bx_t\in \SE{2}$ and a local scale $s\in \mathbb{R}$ at time $t$, given the history of odometry measurements $\bu_{1:t}$ and observations $\bz_{1:t}$.
Formally, this is equivalent to recursively estimate the following posterior:
\begin{eqnarray}
\lefteqn{  p(\bx_t, s_t\mid \bz_{1:t}, \bu_{1:t}, m) = \eta p(\bz_t \mid \bx_t, m)}\cdot\nonumber\\
&& \int_{\bx_{t-1},s_{t-1}} \hspace{-5ex} p(s_t \mid s_{t-1}) p(\bx_t \mid \bx_{t-1}, s_{t-1}, \bu_{t})\cdot\nonumber\\
&& \hspace{-1ex} p(\bx_{t-1}, s_{t-1} \mid  \bz_{1:t-1}, \bu_{1:t-1}, m)\,\mathrm{d}\bx_{t-1}\mathrm{d}s_{t-1},
\label{eq:localization}
\end{eqnarray}
where $\eta$ is a normalization factor and $m$ is the hand-drawn map.
The motion model~$p(\bx_t \mid \bx_{t-1}, s_{t-1}, u_{t})$ denotes the probability that the robot ends up in state~$\bx_t$
given it executes the odometry readings~$\bu_{t}$ in state~$\bx_{t-1}$.
The distribution $p(s_t\mid s_{t-1})$ represents the Brownian motion of the scale parameter and
the observation model $p(\bz_t \mid \bx_t, s_t, m)$ denotes the likelihood of making the observation~$\bz_t$ given the robot's pose~$\bx_t$, the local scale $s_t$, and the map $m$.

Following the MCL approach, we approximate the distribution as a weighted sum of Dirac delta functions.
The recursive estimation is performed using the sequential importance resampling algorithm~\cite{doucet2001sequential}.
For the proposal, we sample the pose and the scale from the motion model and the Brownian process, respectively.
Under the chosen proposal distribution, we compute the weight of the particle according to the observation model and the recursive term.
The particle set is then resampled, at each iteration, according to the weight of the particles.

To compute the number of particles needed, one can use the KLD sampling approach of Fox~\cite{fox03ijrr}.
The algorithm estimates a discrete approximation of the target distribution using the weighted set of particles.
During resampling, it computes the Kullback-Leibler divergence each time a new particle is resampled and stops the process when the
divergence is below a confidence level.

\subsection{Proposal Distribution}

The purpose of the proposal distribution is to provide a mean for sampling a new set of particles given the current one.
In the original MCL algorithm, the proposal distribution is the robot motion model. In our work, we need to provide a proposal distribution for both the robot position $\bx$ and the local scale $s$.
We modified the original motion model describe in the MCL algorithm to project the motion of the robot in the image coordinates.
Let $\bx_t^i$ and $s_t^i$ the pose and scale associated with the $i$-particle at time $t$ and $\bu_t$ the incremental odometry measurement.
The new pose of the particle is computed as follow
\begin{equation}
 \bx_{t+1}^i = \bx_t^i \oplus \mathbf{S}^{-1}(\bu_t\oplus\hat\be)  \qquad \mathbf{S} = \left[\begin{array}{ccc}s_t^i & 0 & 0\\ 0 & s_t^i & 0\\0 & 0 & 1\end{array}\right],
\end{equation}
where $\bu_t$ represents the odometry reading and $\hat\be$ is a sample from the noise term.
We sample $\hat\be = \left[\begin{array}{cc}\bq_t^i &\theta_t^i\end{array}\right]^T$ from the normal distribution and a wrapped normal distribution, respectively for the translation $\bq_t^i$ and the rotational $\theta_t^i$ part.
\begin{eqnarray}
\bq &\sim& \mathcal{N}(\mathbf{0}, \Sigma_\bq)\\
\theta_t &\sim & \mathcal{WN}(0, \sigma_\theta^2),
\end{eqnarray}
where $\Sigma_\bq$ and $\sigma_\theta$ are the covariance matrix for the translation and the standard deviation for the rotation.
The scale follows a Brownian motion and is sampled according to following process
\begin{equation}
 s_{t+1}^i = s_t^i + \epsilon^i \qquad \qquad \epsilon \sim \mathcal{N}(0,\sigma_s^2),\label{eq:scale_motion}
\end{equation}
where $\epsilon^i$ is a sample from a normal distribution and $\sigma_s$ is the standard deviation of the scale increment.
Note that we include a standard deviation term in the Wiener process. This is to account for smaller variations than its original formulation.
One can formulate \eqref{eq:scale_motion} using the standard formulation of the Wiener process by including an additional scaling term to the $\epsilon^i$.

Intuitively, given the pose of a particle and its estimated scale, we first sample a zero mean, $\SE{2}$ transformation from the odometry noise and perturb the odometry measurement accordingly.
We then project the perturbed odometry on the hand-drawn map and apply the projected transformation to the robot pose. The scale sampling has been chosen to be locally close to the scale of the previous step, but still able to explore the whole space. The Brownian motion was a natural choice for that, given its statistical properties.

\subsection{Observation Model}

After we sample the particles according to the proposal distribution, we follow the importance sampling principle and compute the weights of the particle set.
With our choice of proposal, the weight of each particle must be proportional to the observation model $p(\bz \mid \bx, s, m)$, where we omitted the time index for notational simplicity.
Intuitively, this model describes the likelihood of the measurement $\bz$, given the hand-drawn map $m$, the local scale $s$ and the pose of the robot in the image coordinates $\bx$.
In this work, we consider 2D laser range finders as sensors. The measurements $\bz = [z_1,\cdots,z_K]^T$ consist of a set of range values along a set of corresponding directions $\mathbf{a} = [\alpha_1, \cdots, \alpha_K]^T$. We employ the beam based model~\cite{thrun2005probabilistic} and modify it to project the observations in the hand-drawn map.

Formally, let $z_i$ be the $i$-th range measurement along the angle $\alpha_i$. Let us trace a ray on the map $m$ from the robot pose $\bx$ to the closest obstacle in the map and be $\hat{z}$ the measured distance. The original formulation of the beam based model considers $\hat{z}$ as being expressed in world coordinates and describes the measurement distribution as a mixture of four components
\begin{equation}
 p(z_i \mid \hat{z_i}) = \left[\arraycolsep=3pt \begin{array}{c}
			      w_\mathrm{hit} \\ w_\mathrm{dyn} \\ w_\mathrm{max} \\ w_\mathrm{rnd}
			      \end{array}\right]^T \left[\begin{array}{c}
				    f_\mathrm{hit}(z_i, \hat{z}_i)\\
				    f_\mathrm{dyn}(z_i, \hat{z}_i)\\
				    f_\mathrm{max}(z_i, \hat{z}_i)\\
				    f_\mathrm{rnd}(z_i, \hat{z}_i)
			      \end{array}\right],
\end{equation}
where $f_\mathrm{hit}$ models the measurement noise, $f_\mathrm{dyn}$ models the unexpected obstacles not present in the map, $f_\mathrm{max}$ models the sensor maximum range, and $f_\mathrm{rnd}$ models a uniformly distributed random measurement.
The functions are defined as following
\begin{eqnarray}
 f_\mathrm{hit}(z, \hat{z}) &=& \mathcal{N}(z;\hat{z},\sigma_z)\\
 f_\mathrm{dyn}(z, \hat{z}) &=& \mathrm{TEXP}(z;\lambda,\hat{z})\\
 f_\mathrm{max}(z, \hat{z}) &=& \mathcal{U}(z; 0, z_\mathrm{max})\\
 f_\mathrm{rnd}(z, \hat{z}) &=& \mathcal{U}(z; z_\mathrm{max} - \delta, z_\mathrm{max} + \delta).
\end{eqnarray}
Here, $\mathrm{TEXP}(x;\lambda,a)$ denotes a truncated exponential distribution with parameter $\lambda$ and support between $0$ and $a$. $\mathcal{U}(x;b,c)$ denotes a uniform distribution between $b$ and $c$ and $\delta$ is a window parameter.
To account for the deformations, we need to project the real measurements coming from the sensor to the image coordinate frame by applying the estimated transformation.
In our case, this entails to scale all the ranges according to estimated scale $s$. The resulting observation model is
\begin{equation}
 p(z_i \mid s, \hat{z}_i) = p(\frac{z_i}{s} \mid \hat{z_i})
\end{equation}

All the parameters of the model have been learned from real data.
To collect the data for the learning phase, we positioned the robot at fixed locations in the environment and draw few different sketches for each location.
Then, for each sketch and location, we performed grid search to find the best scale. Given the scale, we computed the maximum-likelihood estimate of the parameters, following the approach described in~\cite{thrun2005probabilistic}.

\section{Experimental Results}

\begin{figure}
\begin{center}
\includegraphics[height=0.15\textwidth]{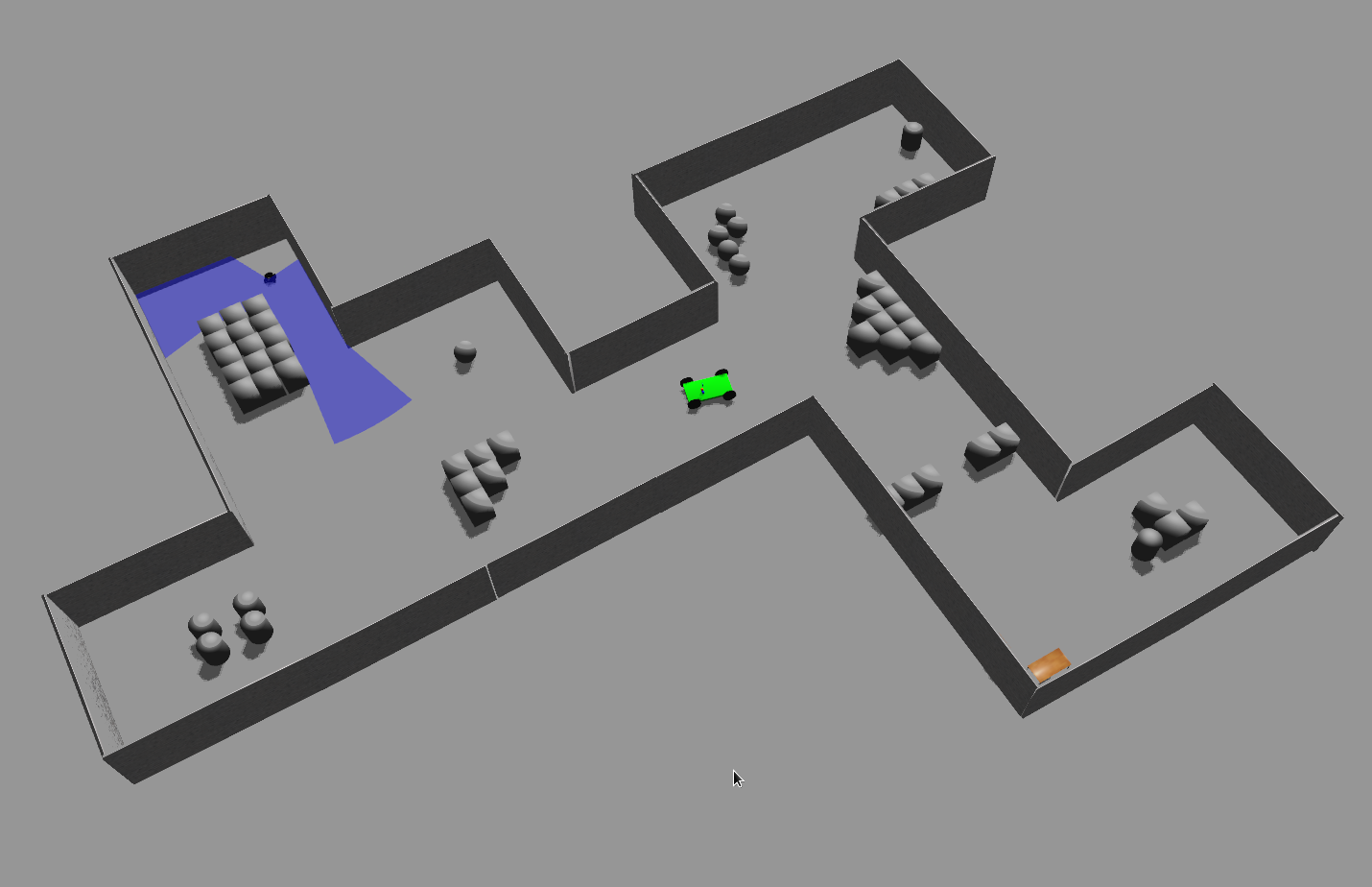}
\includegraphics[height=0.15\textwidth]{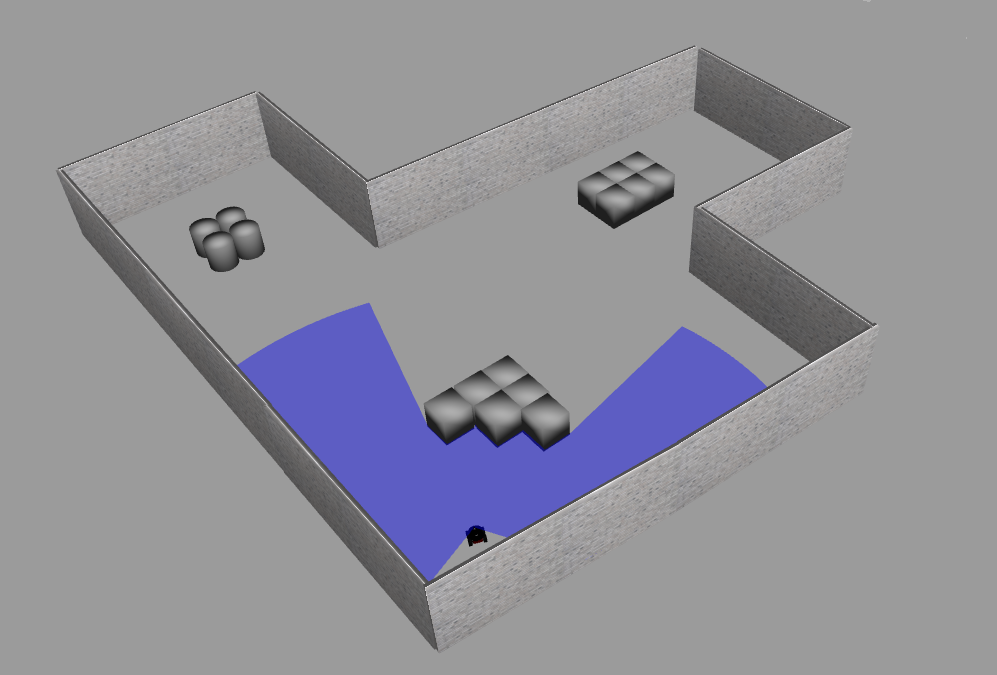}
\end{center}
  \caption{Simulated Apartment (left) and Room (right) environment in Gazebo. The area covered by the simulated laser scanner is shown in blue.}
\label{fig:simpleMap2_3D}
\end{figure}

We evaluated our approach in both simulated and real environments.
For the simulation, we used the Gazebo simulator available in ROS.
We created two virtual environments, resembling a simulated room (Room) and a simulated apartment (Apartment).
Figure~\ref{fig:simpleMap2_3D} depicts the two simulated environments together with a simulated robot.
In the real world experiment, we tested our algorithm in our building (FR079), whose map and a full dataset is publicly available.
We chose this dataset for two reasons.
First it contains very challenging aspects of localization algorithms, given the presence of many, similarly looking rooms.
Second, similar data is publicly available online, and other researchers can use that to replicate our results.
Since not everyone has the same \emph{artistic capabilities}, we asked nine people to walk in the environment and draw a sketch. Figure~\ref{fig:ais_9} shows all the nine sketches together.

We use the same parameters for all our experiments and the simulated robot.
For the proposal distribution, we have $\Sigma_\bq=0.1 I$ as covariance matrix for the translational component, $\sigma_\theta=0.05$ for the rotational noise, and $\sigma_s = 0.1$ for the scale noise. With respect to the observation model, we set $\sigma_z=0.1$ according to our sensor specification, and we estimated the rest of the parameters from data.
The resulting estimated values are $\lambda = 0.1$ for the exponential distribution, $\delta = 0.01$ for the max range, and $w_\mathrm{hit}=0.005$, $w_\mathrm{dyn}=0.5$, $w_\mathrm{max}=0.3$, $w_\mathrm{rnd}=0.4$ for the mixture weights. We also subsampled the range measurements to only 10 beams per scan, to compensate for the independence assumption of the laser beams.
We intialized the filter by drawing the particles uniformly over a region of the map drawn by the user, for the position, uniformly between $-\pi$ and $\pi$ for the orientation, and uniformly between 0.01 and 1 for the scale. The square region was about the size of a room, simulating a plausible initial guess from the user.

\subsection{Simulated Experiments}

For the simulation, we used the Room environment as a proof of concept scenario.
We let the robot start in the lower right corner of the room and performed 4 different paths.
We simulated a weak prior on the initial robot position by sampling the particles uniformly in a square of $150\times150$ centered at the robot position.
We obtained a success rate of 100\% in localizing the robot. Some videos of the whole experiment are available on Youtube\footnote{\url{https://www.youtube.com/playlist?list=PL2DAq2wc_lgJnSTYusQjeck-gc3fc2jtP}}.

\begin{figure}
\begin{center}
\includegraphics[width=0.22\textwidth]{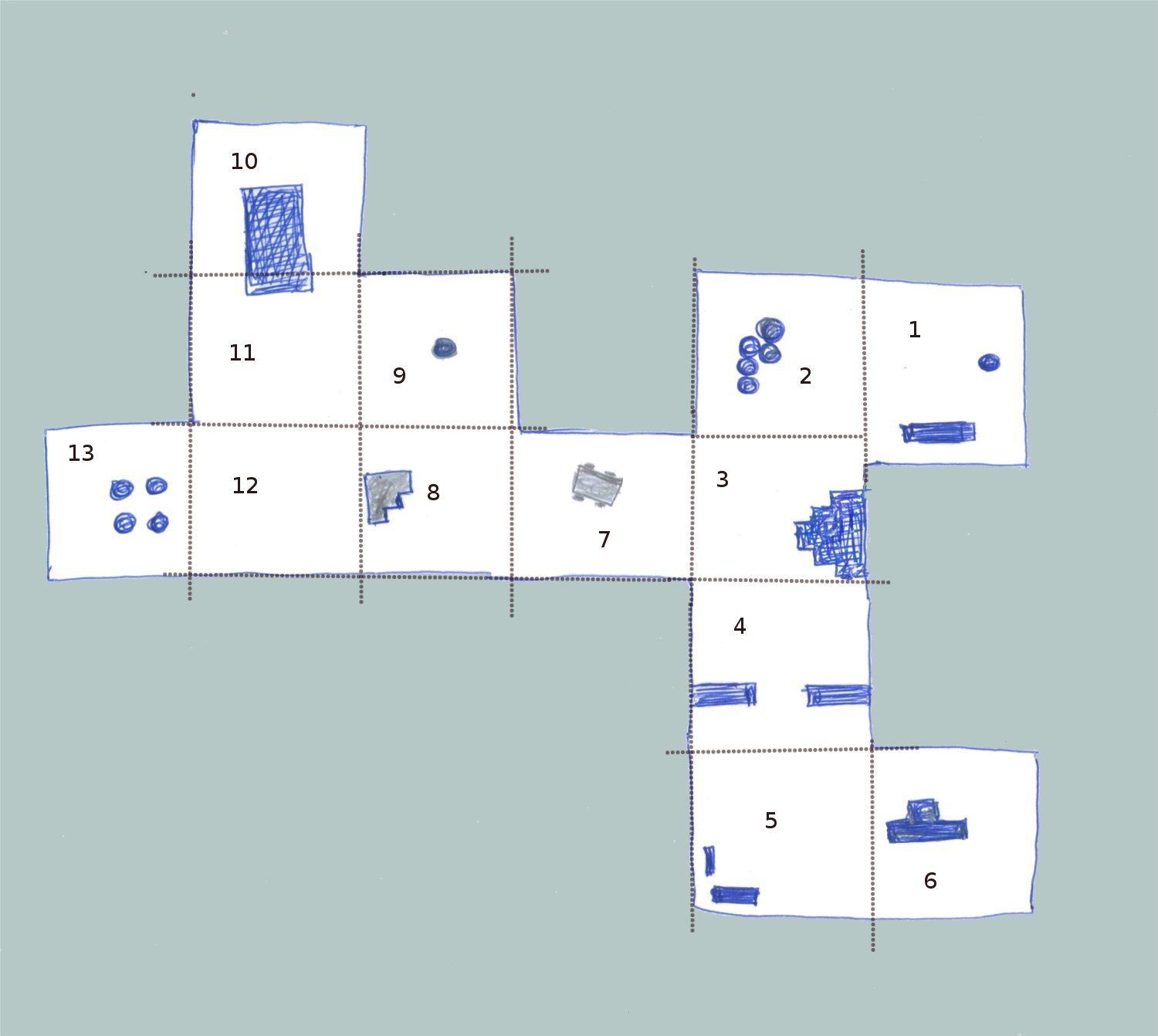}
\includegraphics[width=0.238\textwidth]{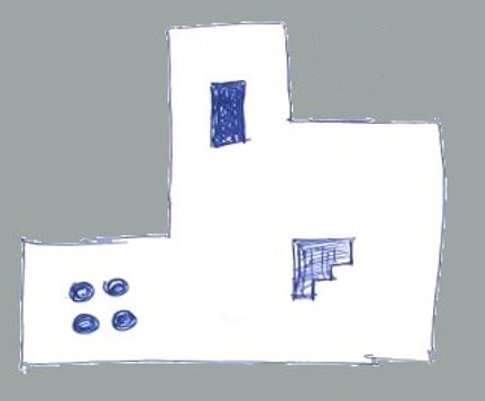}
\end{center}
  \caption{Hand-drawn map of the Apartment (left) and the Room (right) environment. The dashed squares represents the rooms we used in our experiment.}
\label{fig:simpleMap2_3D_regions}
\end{figure}

For the second experiment in the Apartment environment, we simulated a series of navigation tasks, where the robot starts in a certain location and is instructed to reach a specific room.
We believe this is the natural application of our approach, where the user sketches the map, a rough location of the robot and ask the robot to reach a particular showing it on the map.
We set up our experiments in the following way.
After we draw a sketch of the environment, we subdivided it into small square, each representing a room to be reached.
We then randomly generated 10 different sequences navigation tasks, in the form of go from room A to room B.
For each sequence, we performed 10 runs of our localization algorithm with different random seeds.
We considered a sequence as a success if  the robot, at the end of the trajectory, is localized in the correct room.

\begin{figure}
\begin{center}
\includegraphics[width=0.35\textwidth]{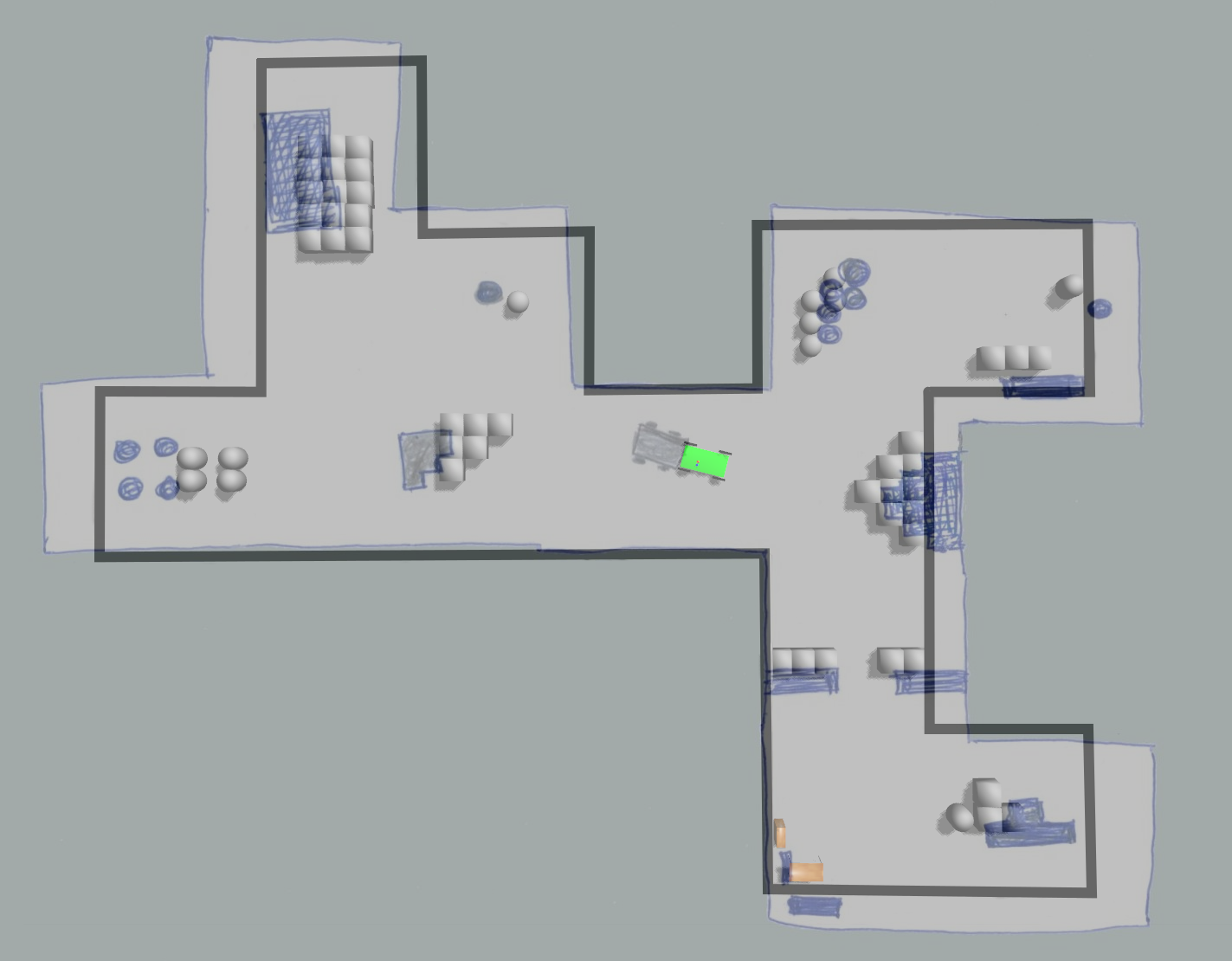}
\end{center}
  \caption{Overlay of the hand-drawn map over the Apartment environment. The hand-drawn map has been manually aligned. Note the non uniform scaling and the distortions present.}
\label{fig:simpleMap2_top_with_sketch}
\end{figure}

Figure~\ref{fig:simpleMap2_3D_regions} shows the hand-drawn map used in this experiments, together with our subdivision.
To understand the differences between the real map and the hand-drawn one, Figure~\ref{fig:simpleMap2_top_with_sketch} shows an overlay of the two, where we manually rotate, scaled and aligned the two maps. Even under manual alignment, one can see that the scaling of the sketch is not uniform with respect to the real map and that many distortions are present.
Table~\ref{tab:table_simpleMap2} shows the results of the experiment, together with the sequences we used.
Our approach has an overall success rate of 93\%. We only had a few failures in the paths from 10 to 6 and 13 to 6.
Note that this are the most challenging paths, since are the longest and traverse the whole map.
We also had some problems from 13 to 10. This was due to the ambiguity in the two corners, where the robot was mistakenly localized in the wrong one.

\begin{table}
\centering
\begin{tabular}{@{}c|c@{}}
\toprule
room a $\rightarrow$ b & Chance of Success \\ \midrule
1 $\rightarrow$ 6    & 100\%             \\
1 $\rightarrow$ 10    & 100\%             \\
6 $\rightarrow$ 1      & 100\%             \\
6 $\rightarrow$ 10     & 100\%             \\
8 $\rightarrow$ 1     & 100\%             \\
8 $\rightarrow$ 6     & 100\%             \\
10 $\rightarrow$ 1      & 100\%             \\
10 $\rightarrow$ 6     & 70\%             \\
13 $\rightarrow$ 6    & 80\%              \\
13 $\rightarrow$ 10     & 80\%              \\ \midrule
Total                  & 93\%
\end{tabular}
\caption{Success rate for the Apartment environment}
\label{tab:table_simpleMap2}
\end{table}

All the trajectories for this experiment are publicly available on youtube\footnote{\url{https://www.youtube.com/playlist?list=PL2DAq2wc_lgI4x4TK1PQOsovRKk-99OwS}}.

\subsection{Real World Experiments}

\begin{figure*}
\centering
{\renewcommand{\arraystretch}{1.3}
\begin{tabular}{MMM}
\includegraphics[height=0.125\textwidth]{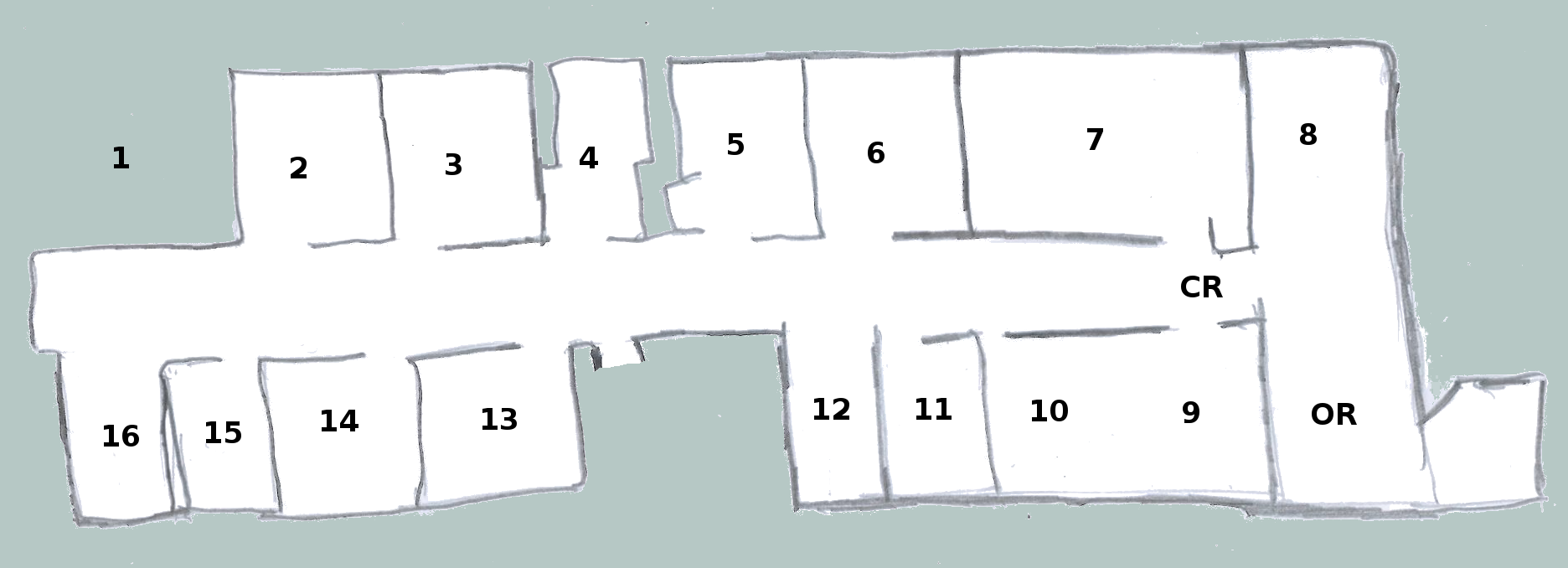} &
\includegraphics[height=0.125\textwidth]{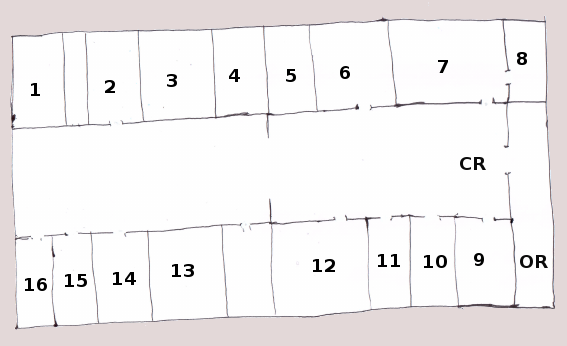} &
\includegraphics[height=0.125\textwidth]{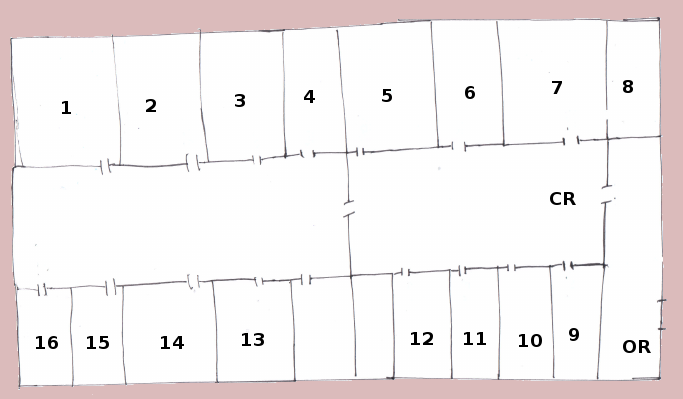} \\
ais-0 & ais-1 & ais-2 \\
\includegraphics[height=0.125\textwidth]{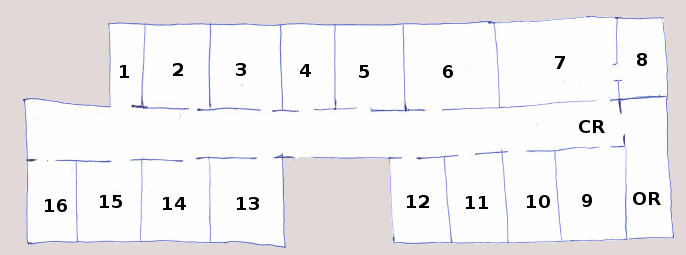} &
\includegraphics[height=0.125\textwidth]{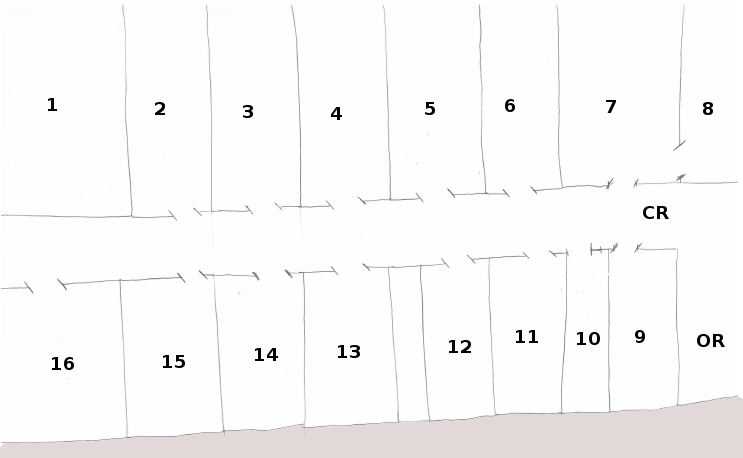} &
\includegraphics[height=0.125\textwidth]{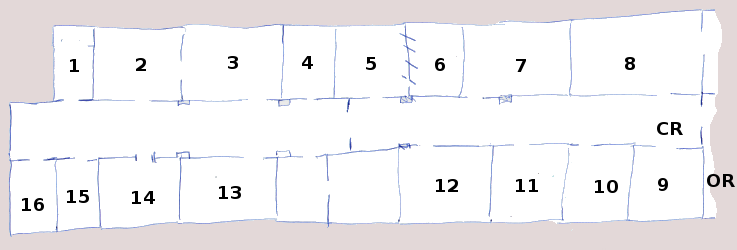} \\
ais-3 & ais-4 & ais-5 \\
\includegraphics[height=0.125\textwidth]{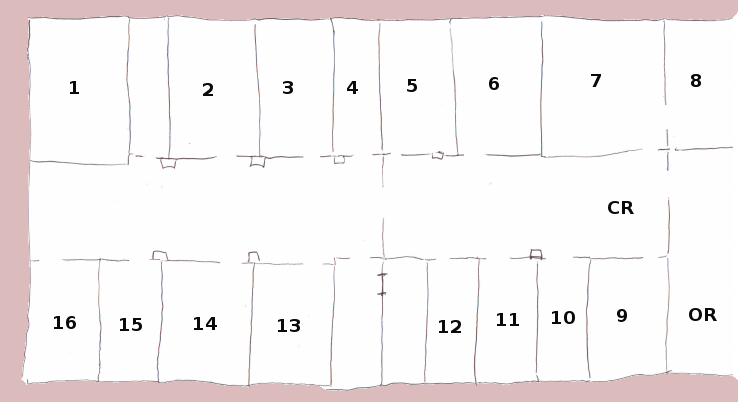} &
\includegraphics[height=0.125\textwidth]{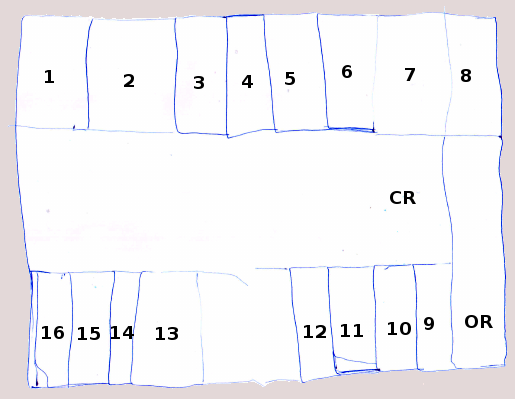} &
\includegraphics[height=0.125\textwidth]{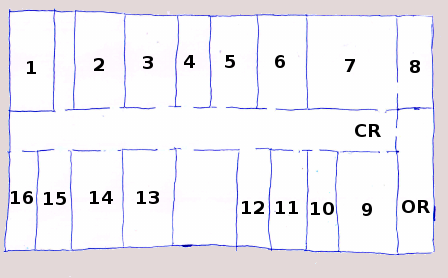} \\
ais-6 & ais-7 & ais-8
\end{tabular}
}
\caption{Sketch maps for the FR079 environment used for the experiments.}
\label{fig:ais_9}
\end{figure*}

\begin{figure}
\centering
\includegraphics[width=0.4\textwidth]{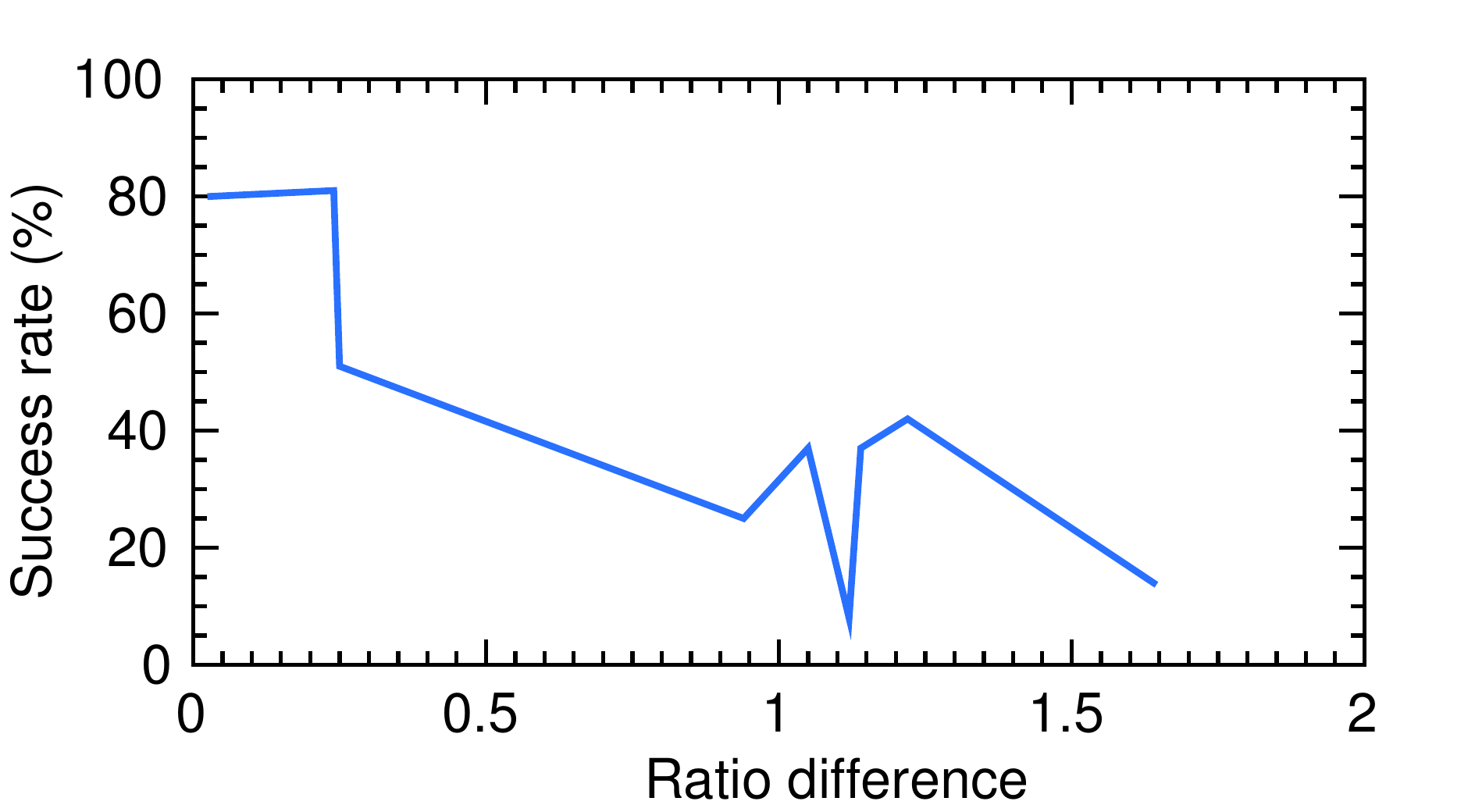}
\caption{Percentage of FR079 runs successfully localized.}
\label{fig:error_perc}
\end{figure}

Figure~\ref{fig:ais_9} shows the hand-drawn maps of Building 079 used for this experiment. The numbers in the figure represent the different rooms we identified in the environment.
In a way similar to the simulated experiments, we randomly generate 7 navigation sequences and, for each sequence, we performed 10 runs of our localization algorithm with different random seeds.
Table~\ref{tab:big_table} shows the localization results with the real data for each run.
The ratio difference denotes the absolute difference between the ratio (length/width) of the original occupancy grid map and each hand-drawn map.
Figure~\ref{fig:error_perc} illustrates the success rate as a function of the difference in ratios.
We see that localization has a high success rate, almost 80\%, when the difference in the ratio of the hand-drawn map is relatively low.
The success rate of localization decline, when this difference increases.
The table shows the highest failure in the test run from room 9 to 12.
Room number 9 is fully occupied with furniture that heavily distorted the image of the walls in the laser scan, therefore, the robot was not able to localize properly in the beginning.
The robot randomly localized itself in any of the other rooms looking alike.
The lowest successful rate was obtained when using the map No. 2.
The user has drawn the doors in an unusual way that the robot can not recognize the entrance and exit properly.
The localization results for
AIS map-0\footnote{\url{https://www.youtube.com/playlist?list=PL2DAq2wc_lgJu9ftPMaozs0fl9kwrkXCZ}}
and map-3\footnote{\url{https://www.youtube.com/playlist?list=PL2DAq2wc_lgJdzfJJDNUNvom0UWCU3Mz4}}
and is publicly available on youtube.

In addition to FR079, we also tested our method on the Intel dataset. The videos
from the Intel dataset are publicly available on
youtube\footnote{\url{https://www.youtube.com/watch?v=uQhK19jpa2I&index=2&list=PL2DAq2wc_lgI9JDwGfDLvZVds_vwpk3OS}}.

\begin{table*}[t]
\centering
\begin{tabular}{l | r r r r r r r r r }
room-to-room    & ais-3 & ais-0 & ais-5 & ais-6 & ais-1 & ais-2 & ais-8 & ais-4 & ais-7 \\
\hline
ratio difference      & 0.03  & 0.24  & 0.25  & 0.94  & 1.05  & 1.12  & 1.14  & 1.22  & 1.64  \\
\hline
15 to 16        & 100   & 100   & 100   & 80    & 100   & 60    & 100   & 20    & 40    \\
14 to 13        & 80    & 100   & 0     & 40    & 60    & 0     & 60    & 100   & 0     \\
4 to 6          & 100   & 100   & 100   & 0     & 20    & 0     & 0     & 80    & 0     \\
9 to 12         & 0     & 0     & 60    & 0     & 0     & 0     & 0     & 0     & 0     \\
5 to 11         & 100   & 100   & 0     & 0     & 0     & 0     & 0     & 0     & 20    \\
CR to 12        & 80    & 100   & 100   & 0     & 0     & 0     & 100   & 0     & 0     \\
11 to OR        & 100   & 70    & 0     & 60    & 80    & 0     & 0     & 100   & 40    \\
\hline
success rate \% & 80    & 81    & 51    & 25    & 37    & 8     & 37    & 42    & 14    \\
\end{tabular}
\caption{Results of the real world experiments.}
\label{tab:big_table}
\end{table*}

\section{Discussion}
We believe our approach will also work with blueprints of the environment, given their metrical accuracy.

\section{Conclusions}

In this paper, we addressed the problem of robot localization when no
accurate map of the environment is available and the robot has to
solely rely on a hand-drawn map sketched by the user.  To do so, we
extended the classical Monte Carlo localization algorithm in two ways.
First, we propose to localize with respect to the image coordinate
frame.  Second, we track, together with the pose of the robot, a local
deformation of the hand-drawn map.  Since no metric information is
available on the hand-drawn map, we propose to evaluate the
localization in terms of coarse localization at the level of rooms of
the environment.  We evaluated our approach in both simulated and real
environments and achieved a correct localization, up to the room
level, of about 80\% of the cases when the ratio of the sketch map resembles the real environment.
We believe this is a starting point that addresses a very challenging problem with potential applications.
In future, we plan to extend our approach to incorporate more sophisticated distortion models and employ it for navigation purposes.

\bibliographystyle{plainnat}
{\footnotesize
\bibliography{bahram15icra}
}

\end{document}